\documentclass[lettersize,journal]{IEEEtran}
\usepackage{graphicx}
\usepackage{cite}
\usepackage{picinpar}
\usepackage{amsmath}
\usepackage{url}
\usepackage{flushend}
\usepackage[latin1]{inputenc}
\usepackage{colortbl}
\usepackage{soul}
\usepackage{multirow}
\usepackage{pifont}
\usepackage{amsthm}
\usepackage[caption=false,font=normalsize,labelfont=sf,textfont=sf]{subfig}
\usepackage{alltt}
\usepackage{stfloats}
\usepackage[hidelinks]{hyperref}
\usepackage{enumerate}
\usepackage{siunitx}
\usepackage{verbatim}
\usepackage{url}
\usepackage{array}
\usepackage{epstopdf}
\usepackage{pbox}
\usepackage{amssymb,amsfonts}
\usepackage{mathdots}
\usepackage{yhmath}
\usepackage{subcaption}
\usepackage{textcomp}
\usepackage{calligra}
\usepackage{algorithm}
\usepackage{algorithmicx}
\usepackage{algpseudocode}
\usepackage{etoolbox}
\usepackage{booktabs}
\usepackage{adjustbox}
\usepackage{gensymb}
\usepackage{threeparttable}
\usepackage{color}
\usepackage[normalem]{ulem}
\usepackage{nomencl}
\usepackage{bm}
\usepackage{mdwlist,lipsum}
\usepackage[table,xcdraw]{xcolor}
\definecolor{darkgreen}{RGB}{0,170,0} % 暗绿色的RGB表示
\definecolor{darkred}{RGB}{170,0,0} % 暗红色的RGB表示
\definecolor{deepurple}{RGB}{191, 64, 191}
\def\BibTeX{{\rm B\kern-.05em{\sc i\kern-.025em b}\kern-.08em
    T\kern-.1667em\lower.7ex\hbox{E}\kern-.125emX}}
\usepackage{balance}
\usepackage{microtype}

\newtheorem{remark}{Remark}
\newtheorem{lemma}{Lemma}

\begin{document}
\title{SwordRiding: A Unified Navigation Framework for Quadrotors in Unknown Complex Environments via Online Guiding Vector Fields}

\author{
    \vskip 1em
    Xuchen Liu$^{*}$, Ruocheng Li$^{*}$, Bin Xin, Weijia Yao, Qigeng Duan, Jinqiang Cui, Ben M. Chen and Jie Chen
    \thanks{
    $^{*}$Xuchen Liu and Ruocheng Li contributed equally to this work.  
    }
    \thanks{Xuchen Liu and Jinqiang Cui are with Pengcheng Laboratory, Shenzhen, Guangdong, China (e-mail: \{liuxch,cuijq\}@pcl.ac.cn). }
    \thanks{Ruocheng Li and Bin Xin are with the School of Automation, Beijing Institute of Technology, Beijing, China (e-mail: \{ruochengli,brucebin\}@bit.edu.cn). }
    \thanks{Weijia Yao is with the School of Artificial Intelligence and Robotics, Hunan University. (e-mail: wjyao@hnu.edu.cn)}
    \thanks{Chen Jie is with the Department of Control Science and Engineering, Harbin Institute of Technology, and the National Key Laboratory of Autonomous Intelligent Unmanned Systems. (e-mail:  chenjie@bit.edu.cn)}
    \thanks{Qigeng Duan and Ben M. Chen are with the Department of Mechanical and Automation Engineering, the Chinese University of Hong Kong, Hong Kong, China. (e-mail: \{qigenduan,bmchen\}@cuhk.edu.hk)} 
}

\maketitle
	
\begin{abstract}
Although quadrotor navigation has achieved high performance in trajectory planning and control, real-time adaptability in unknown complex environments remains a core challenge. This difficulty mainly arises because most existing planning frameworks operate in an open-loop manner, making it hard to cope with environmental uncertainties such as wind disturbances or external perturbations. 
This paper presents a unified real-time navigation framework for quadrotors in unknown complex environments, based on the online construction of guiding vector fields (GVFs) from discrete reference path points.
In the framework, onboard perception modules build a Euclidean Signed Distance Field (ESDF) representation of the environment, which enables obstacle awareness and path distance evaluation.
The system first generates discrete, collision-free path points using a global planner, and then parameterizes them via uniform B-splines to produce a smooth and physically feasible reference trajectory.
An adaptive GVF is then synthesized from the ESDF and the optimized B-spline trajectory. Unlike conventional approaches, the method adopts a \emph{closed-loop} navigation paradigm, which significantly enhances robustness under external disturbances.
Compared with conventional GVF methods, the proposed approach directly accommodates discretized paths and maintains compatibility with standard planning algorithms. 
Extensive simulations and real-world experiments demonstrate improved robustness against external disturbances and superior real-time performance.\footnote{Video: \url{https://www.youtube.com/watch?v=tKYCg266c4o}}
\end{abstract}


\markboth{}%
{}

\definecolor{limegreen}{rgb}{0.2, 0.8, 0.2}
\definecolor{forestgreen}{rgb}{0.13, 0.55, 0.13}
\definecolor{greenhtml}{rgb}{0.0, 0.5, 0.0}

% \section*{Nomenclature}
% \begin{basedescript}{\desclabelstyle{\pushlabel}\desclabelwidth{8em}}
%     \setlength{\itemsep}{3pt}

%     % === B-spline Parameters ===
%     \item[$p$] Degree of the B-spline
%     \item[$N, M$] Number of control points minus one ($N$), and knot vector length index ($M=N+p+1$)
%     \item[$\mathbf{C}_i$] $i$-th B-spline control point in $\mathbb{R}^3$
%     \item[$u_i, \Delta u$] $i$-th knot vector element and uniform knot spacing
%     \item[$\beta, t_{M-p}, \Delta t$] Time scaling factor between $u$ and $t$, total trajectory time, and time interval between control points
    
%     % === Optimization Weights ===
%     \item[$\lambda_s, \lambda_c$] Weights for smoothness and collision cost terms

%     % === Collision Parameters ===
%     \item[$d(\mathbf{C}_i), d_{\mathrm{thr}}$] ESDF distance from $\mathbf{C}_i$ to nearest obstacle, and collision distance threshold

%     % === GVF Parameters ===
%     \item[$K_1, K_2$] Tangential and normal gains in GVF

% \end{basedescript}

\section{Introduction}
\label{Introduction}
\IEEEPARstart{A}{utonomous} navigation has become a central topic of interest in the field of robotics\cite{11155197}. In a wide range of real-world applications, such as disaster response, agricultural automation, traffic monitoring, and package delivery, navigation serves as a core component of robotic systems. Some existing works, such as \cite{10649014,zhou2020ego,zhou2021raptor,zhu2019chance,tordesillas2021mader}, have investigated navigation in static environments as well as in dynamic environments with unknown obstacles, and have achieved promising results. In these works, the fundamental idea of navigation originates from trajectory planning. That is, given structured environmental data (typically in the form of an occupancy grid map), the robot searches for a feasible path primitive based on optimization principles, and subsequently refines the obtained path through further optimization. The resulting primitive, which satisfies constraints such as safety and dynamic feasibility, is referred to as a locally optimal trajectory. The reference trajectory is ultimately discretized into samples and passed to the low-level controller for execution. In effect, the high-level planner generates desired paths, and the low-level controller tracks them. This strategy follows an open-loop paradigm, which is usually adequate for many practical tasks under predictable conditions\cite{doi:10.1142/S230138502450033X,xu2022dpmpc,li2021collision}.

Nevertheless, in more general or unpredictable operating conditions, this open-loop paradigm becomes inadequate. For example, an aerial-aquatic amphibious quadrotor~\cite{liu2023tj,liu2024tj} cannot rely on existing open-loop navigation methods, as the drag forces in water and air differ fundamentally. Applying the same open-loop approach would prevent the robot from following the reference trajectory in water, ultimately leading to navigation failure. Even in aerial environments, external factors such as wind disturbances or human interventions may occur, and these can still cause planning failures. Although some researchers have incorporated disturbance rejection against wind at the low-level control layer \cite{liu2020affine,bisheban2018geometric}, the problem remains unsolved because such disturbances are neglected at the navigation level. While low-level controllers can handle small disturbances, their capability is inherently limited under sustained or unpredictable perturbations. Hence, a \emph{closed-loop} navigation mechanism is essential to continuously correct trajectory deviations and ensure robustness.

\begin{figure}[t]
\centerline{
\includegraphics[width=0.9\linewidth]{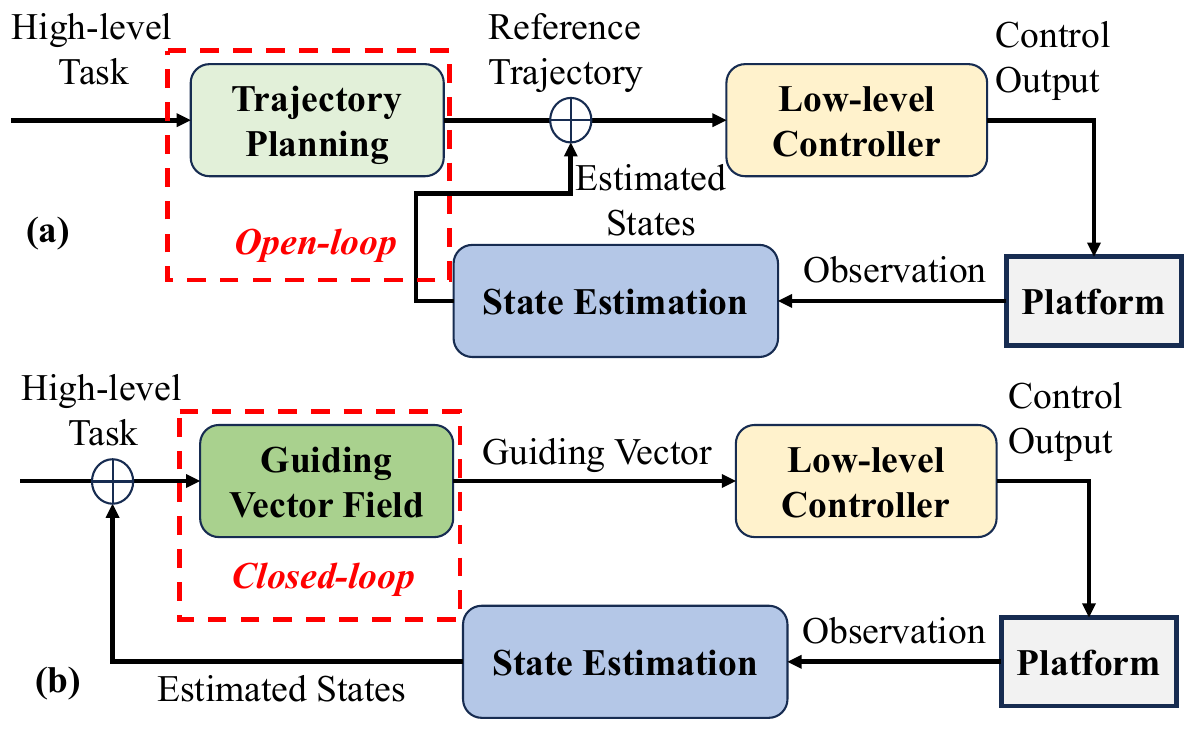}}
\caption{
(a) Open-loop architecture based on trajectory planning.
(b) Closed-loop architecture based on guiding vector field.
}
\label{f1}
\end{figure}

Another typical class of methods for navigation is based on \emph{guiding vector fields}, which has been extensively studied \cite{liang2016combined,rezende2018robust,goncalves2010vector,10639185,11181562,yao2018robotic}. The idea of a \emph{guiding vector field} is that, given a reference path, one can compute a control sequence from any point in the space based on the guiding vectors, such that the robot moves towards along these guiding directions and asymptotically converges to the reference path. Unlike trajectory planning, a \emph{guiding vector field} provides a \emph{closed-loop} construction. In other words, the control input of a robot depends on its current position: if it is already on the reference path, the input will keep it moving along the path, whereas if it is off the path, the guiding vector will drive it back toward the reference. Fig.~\ref{f1} illustrates a comparison between trajectory planning and the \emph{guiding vector field} approach. In trajectory planning, the initial condition of the reference trajectory is determined by the feedforward states rather than the instantaneous feedback, in order to ensure trajectory smoothness and to provide the controller with the necessary velocity and acceleration feedforward terms. If state feedback were directly used as the starting point, the replanned trajectory would suffer from discontinuities or even stagnation. In contrast, the \emph{guiding vector field} approach relies on local feedback to compute control inputs, thereby avoiding such discontinuity issues\cite{yao2020vector}.

For guiding-vector-field-based navigation, Goncalves et al.~\cite{goncalves2010vector} proposed an artificial vector field framework for navigating along time-varying curves in $n$-dimensional space. Rezende et al.~\cite{rezende2021constructive} constructed time-varying vector fields based on nearest-point distances and proved their convergence and robustness. Yao et al.~\cite{yao2021singularity} introduced a singularity-free guiding vector field to guarantee global convergence along complex paths. Li and Calinon~\cite{li2025movement} transformed motion primitives into distance fields and then into dynamical systems to achieve robust motion representation. Wang et al.~\cite{10639185} combined discrete waypoints with time-varying vector fields to achieve online path generation and convergence. Li et al.~\cite{Li24RSS,Li24ICRA} proposed distance-field-based approaches in configuration space, including the use of configuration space distance fields for manipulation planning and a geometric representation of robots to support whole-body manipulation. 

However, guiding vector field methods are difficult to apply independently in complex and unknown environments. The core limitation lies in their reliance on predefined paths represented by implicit functions; in the absence of explicitly represented paths, it is not possible to construct effective convergence and propagation terms, thereby preventing the realization of closed-loop navigation that integrates both path generation and path following. Although existing methods such as \cite{10639185,rezende2021constructive} consider obstacles, they still assume a known obstacle environment. To the best of our knowledge, existing \emph{guiding vector field} approaches address either open environments or simple, fully-known obstacle configurations, while significantly more complex and unknown environments remain largely unexplored.

Motivated by the above facts, in this paper we propose \textbf{SwordRiding}, a navigation algorithm where an incrementally constructed vector field acts as a \emph{guiding sword} to pilot the quadrotor autonomously through complex and unknown environments. First, we address the limitation of conventional guiding-vector-field methods that rely on implicit function representations of the desired path in order to analytically construct guiding vectors. Instead, a trajectory-induced Euclidean distance field is introduced, obtained by discretizing the reference trajectory and computing the corresponding Euclidean distance transform. Based on this idea, the normal component of the guiding vector is constructed from the gradient of the Euclidean distance field, while the tangential component is derived through grid interpolation. This approach fundamentally removes the dependence on analytic functions for constructing guiding vectors in traditional methods, enabling the representation of curves with arbitrary geometries and varying levels of smoothness and curvature complexity. Building on this framework, we employ a multi-objective optimization method to generate reference trajectories using B-spline curves as primitives and defining cost functionals over spline control points to ensure safety and feasibility.

Unlike trajectory planning, which requires the robot to forcibly follow a time-varying reference point, a \emph{guiding vector field} generates a guiding vector based on the robot's current real-time state, and can thus be regarded as a state-feedback closed-loop navigation scheme. This architecture enables the robot to effectively counteract environmental disturbances. For instance, if wind or other interferences cause the robot to deviate from the initial reference state, the vector field generates guiding vectors based on the real-time state to drive the robot back smoothly. During this process, the robot can return to the reference trajectory and continue forward, without exhibiting erratic responses, thereby greatly enhancing system robustness. Moreover, unlike existing vector field methods, our approach can generate vector fields along arbitrary trajectories that are homeomorphic to a $C^1$ manifold, and can be constructed online in an incremental manner. This capability allows the method to effectively adapt to unknown and complex environments, representing a significant improvement over existing vector field approaches. We summarize our main contributions as follows:

1. A novel guiding-vector-field representation is proposed, in which a trajectory-induced Euclidean distance formulation is developed to construct the field directly from discrete reference path points. Unlike conventional GVF methods that depend on analytically or parametrically defined paths, the proposed representation performs a distance transform over the discretized trajectory to overcome the limitation of representing irregular or non-smooth paths. This approach eliminates the dependency on specific path parameterizations and unifies the construction of guiding fields across arbitrary trajectory representations, enabling direct geometric encoding without explicit functional modeling.

2. Based on this representation, a closed-loop navigation framework is established, integrating real-time field updates, adaptive trajectory refinement, and feedback regulation within a single architecture.
The proposed guiding-vector-field-based framework establishes a feedback mechanism that corrects deviations under environmental disturbances, achieving inherent robustness against wind and human-induced perturbations. Extensive simulation and real-world experiments on quadrotor platforms validate the framework's robustness and real-time performance compared with conventional open-loop and fixed-field methods.

The remainder of this paper is organized as follows. Sec.~\ref{Guiding Vector Field Approach} introduces the general guiding vector field approach and discusses its limitations. Sec.~\ref{Navigation Method based on Guiding Vector Field} presents the construction method of the incremental guiding vector field. Sec.~\ref{Results} reports simulation and physical experiments together with a comparative discussion of the results. Finally, Sec.~\ref{Conclusion and Future Work} concludes the paper and outlines future research directions.

\section{Guiding Vector Field Approach}
\label{Guiding Vector Field Approach}
\subsection{Constructing Navigation via Guiding Vector Fields}
\label{Constructing Navigation via Guiding Vector Fields}

We begin by summarizing the essential principles of the guiding vector field (GVF) method for robot navigation, which establishes the geometric foundation for our proposed approach.  

In path-following navigation, the objective is to steer a robot toward and along a desired geometric path without explicit time parameterization. In contrast to trajectory tracking methods, which rely on precise timing and may suffer from instability on underactuated platforms, GVF-based navigation focuses purely on spatial convergence. The path is regarded as a geometric curve embedded in the workspace, and the navigation problem reduces to designing a vector field whose integral curves converge to this curve.  

Let $\mathcal{P} \subset \mathbb{R}^n$ denote the desired path, represented implicitly as the intersection of level sets of smooth scalar fields:  
\begin{equation}
\mathcal{P} = \left\{ \xi \in \mathbb{R}^n \;\middle|\; \phi_i(\xi) = 0, \quad i = 1, \dots, n-1 \right\},
\end{equation}
where each $\phi_i : \mathbb{R}^n \rightarrow \mathbb{R}$ is assumed to be twice continuously differentiable. The guiding vector field $\chi : \mathbb{R}^n \rightarrow \mathbb{R}^n$ is then constructed as  
\begin{equation}
\chi(\xi) = \nabla\!\times\!\phi(\xi) - \sum_{i=1}^{n-1} k_i \, \phi_i(\xi) \nabla \phi_i(\xi),
\end{equation}
where $\nabla\!\times\!\phi(\xi)$ denotes the generalized cross product \cite{yao2021singularity} of the gradients $\nabla \phi_1(\xi), \dots, \nabla \phi_{n-1}(\xi)$, and the constants $k_i > 0$ are control gains. Intuitively, the GVF consists of two orthogonal components. The first is the \emph{propagation term}, $\boldsymbol{t}(\xi)\!:=\nabla\!\times \!\phi(\xi)$, which is tangent to the desired path and promotes motion along it. The second is the \emph{convergence term}, $\boldsymbol{n}(\xi)\!:= -\sum_{i=1}^{n-1} k_i \,\phi_i(\xi)\,\nabla \phi_i(\xi)$, which attracts the system trajectories toward the path. Taken together, these components provide the geometric mechanism that advances the system along the reference path while correcting deviations.

This construction reveals the central idea of GVF-based navigation: by combining the propagation term $\boldsymbol{t}(\xi)$ and the convergence term $\boldsymbol{n}(\xi)$, one obtains a closed-loop guidance law in which the robot's control inputs depend directly on its instantaneous state. As a result, the robot remains on the reference path when aligned with it, while being robustly driven back when deviating from it. Geometrically, the propagation component represents a direction tangent to the path, obtained as the cross product of the gradients of the defining functions; for instance, in $n{=}3$ one has $\boldsymbol{t}\!= \!\nabla \phi_1\!\times\!\nabla \phi_2$, while in $n{=}2$ it reduces to a simple rotation $\boldsymbol{t}\!=\!\mathbf{R}_{\pi/2}\nabla \phi$. In contrast, the convergence component captures the transversal influence of the level sets, driving trajectories back toward the path through a weighted combination of their gradients. Their joint effect is to drive the system forward along the desired path while reducing deviations from the desired path.

For numerical stability, these directions are typically normalized as  
\begin{equation}
\hat{\boldsymbol{t}}(\xi) = \frac{\boldsymbol{t}(\xi)}{\max(\|\boldsymbol{t}(\xi)\|,\varepsilon)}, 
\qquad
\hat{\boldsymbol{n}}(\xi) = \frac{\boldsymbol{n}(\xi)}{\max(\|\boldsymbol{n}(\xi)\|,\varepsilon)}.
\end{equation}

The guiding vector field is then synthesized as  
\begin{equation}
\chi(\xi) = \hat{\boldsymbol{t}}(\xi) + k(\xi)\,\hat{\boldsymbol{n}}(\xi), \qquad k(\xi)>0,
\end{equation}
which simultaneously promotes motion along the path and convergence back to it when deviations occur.  

\subsection{Insights on Tangential and Normal Components}
\label{Insights on Tangential and Normal Components}

Under appropriate assumptions on the regularity of the desired path and the absence of singularities (i.e., points where $\chi(\xi) = 0$), one can show that the integral curves of $\chi$ asymptotically converge to $\mathcal{P}$ \cite{yao2021singularity}. This property makes GVF methods particularly suitable for aerial, ground, and underwater vehicles, where path convergence and alignment are often more critical than precise timing.

Nevertheless, the practical implementation of GVF methods poses several challenges:
\begin{itemize}
\item Constructing $\phi_i$ from arbitrary geometric paths is nontrivial, especially when the path is specified by parameterized curves rather than analytic level sets.
\item Incremental construction of vector fields lacks an effective guiding mechanism, which hinders the application of GVF methods in unknown and complex environments.
\end{itemize}

These difficulties mainly stem from the reliance of classical GVF formulations on implicit functions to represent the desired path. When the path is only expressed in the form of parameterized curves or discrete waypoints, it becomes unclear how to construct the required implicit level sets, and consequently how to obtain the guiding components. If, however, one can directly extract the tangential and normal directions from the given path representation, this difficulty can be circumvented, and the construction of guiding vector fields becomes considerably more general and flexible. The foregoing analysis therefore indicates that the essence of GVF construction lies in obtaining the explicit tangential and normal directions, rather than in relying on implicit analytic functions of the path. For instance, given a smooth curve $\gamma:[0,1]\to\mathbb{R}^n$ with point $\xi=\gamma(s)$, the tangent $\boldsymbol{t}(\xi)=\dot{\gamma}(s)/\|\dot{\gamma}(s)\|$ and the corresponding normal $\boldsymbol{n}(\xi)$ satisfying $\langle \boldsymbol{t}(\xi),\boldsymbol{n}(\xi)\rangle=0$ are sufficient to define a guidance direction of the form $\chi(\xi)=\boldsymbol{t}(\xi)+k(\xi)\boldsymbol{n}(\xi)$. This illustrates that the guiding principle is independent of the specific representation of the path, provided that the necessary geometric primitives are available.

This perspective can be formalized by introducing an operator
\begin{equation}
\Pi:\;\mathcal{C}(\mathbb{R}^n)\;\longrightarrow\;\mathcal{V}(\mathbb{R}^n),
\end{equation}
which assigns to any admissible path $\mathcal{P} \in \mathcal{C}(\mathbb{R}^n)$ the corresponding tangential and normal directions $(\boldsymbol{t}(\xi), \boldsymbol{n}(\xi)) \in \mathcal{V}(\mathbb{R}^n)$ at each point $\xi \in \mathcal{P}$. Here, $\mathcal{C}(\mathbb{R}^n)$ denotes the set of sufficiently smooth paths in $\mathbb{R}^n$, and $\mathcal{V}(\mathbb{R}^n)$ denotes the set of tangent-normal vector pairs. Once such a mapping $\Pi(\mathcal{P}) = (\boldsymbol{t},\boldsymbol{n})$ is available, the guiding vector field follows directly as a structured combination of these components, thereby providing a more general and flexible framework than formulations dependent on implicit path functions. This operator-based viewpoint highlights that the availability of tangential and normal directions alone is sufficient for synthesizing a GVF. The following proposition formalizes this observation and establishes the convergence of the resulting integral curves.

\begin{lemma}
\label{lem:GVF_construction}
Let $\mathcal{P} \subset \mathbb{R}^n$ be a smooth path, and suppose there exists a mapping operator $\Pi$ that assigns to each point $\xi \in \mathcal{P}$ a pair of normalized vectors $(\boldsymbol{t}(\xi),\boldsymbol{n}(\xi))$, with $\boldsymbol{t}(\xi)$ tangential to $\mathcal{P}$ and $\boldsymbol{n}(\xi)$ chosen to be linearly independent from $\boldsymbol{t}(\xi)$. Then one can construct a guiding vector field of the form
\begin{equation}
\chi(\xi) = \boldsymbol{t}(\xi) + k(\xi)\,\boldsymbol{n}(\xi), \qquad k(\xi) > 0,
\end{equation}
 under standard smoothness and Lipschitz continuity assumptions on $\Pi$ and $k(\xi)$, the corresponding integral curves asymptotically converge to $\mathcal{P}$.
\end{lemma}

\noindent\textit{Proof.} 
A detailed proof of this lemma can be found in the supplementary material provided in our repository\footnote{\url{https://github.com/SmartGroupSystems/GVF_close_loop_planning/blob/main/proofs.md}}.

\section{Incremental Guiding Vector Field Navigation}
\label{Navigation Method based on Guiding Vector Field}

\subsection{Incremental Construction of Guiding Vector Fields}
\begin{figure*}[htbp]
\centerline{
\includegraphics[width=0.85\linewidth]{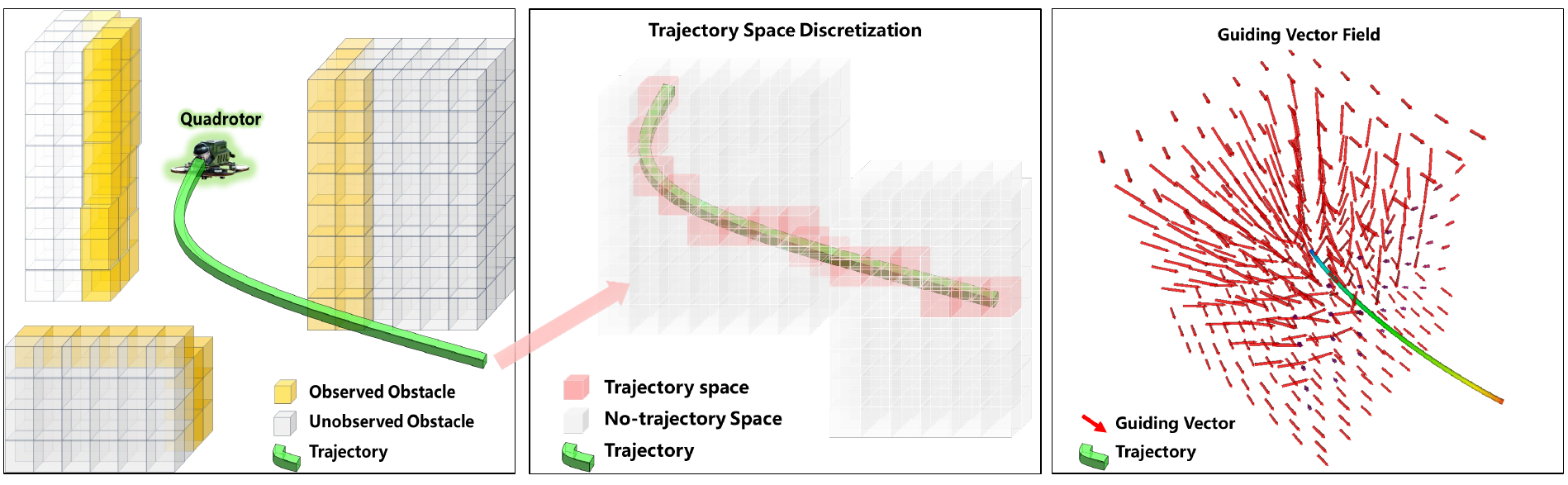}}
    \caption{%
    Generation details of the incremental GVF method.
    \textbf{Left:} a quadrotor plans a feasible reference trajectory through the observed free space while avoiding known obstacles. 
    \textbf{Middle:} the optimized trajectory is discretized into a finite set of path points, forming the trajectory space~$\mathcal{P}$ embedded within the ESDF representation of the environment. 
    \textbf{Right:} an incremental guiding vector field is synthesized around the discrete trajectory; 
    the resulting field provides local flow directions that guide the robot smoothly along the reference path while ensuring obstacle avoidance.}
    \vspace{-0.7\baselineskip} % 
\label{f2}
\end{figure*}
Building upon the general principles of guiding vector fields discussed in the previous section, we now turn to the problem of constructing GVFs in unknown and cluttered environments. As noted earlier, classical GVF formulations typically rely on an analytic description of the reference path, which is often impractical when the environment is partially observed or dynamically changing. To overcome this limitation, we develop an incremental construction method that directly utilizes discrete path points together with a Euclidean Signed Distance Field (ESDF), as illustrated in Fig.~\ref{f2}.

Specifically, consider a robot navigating through an environment populated with partially unknown obstacles. As illustrated in the left panel of Fig.~\ref{f2}, the robot first computes a feasible reference trajectory by solving a global optimization problem that balances collision avoidance, path length, and smoothness. The resulting trajectory provides a safe passage through the currently observed free space and serves as the geometric backbone for subsequent field construction. This trajectory is then discretized into a finite set of path points, as shown in the middle panel of Fig.~\ref{f2}, forming a trajectory space $\mathcal{P}$. To embed this trajectory within the environment representation, each grid cell traversed by the path is associated with the corresponding occupied state, while cells outside the path remain marked as free. 

On this discretized space, we apply the Euclidean distance transform to obtain a signed distance field $U$, defined pointwise by
\begin{equation}
d(\xi) = \min_{P_i \in \mathcal{P}} \|\xi - P_i\|, \qquad \xi \in \mathbb{R}^n,
\end{equation}
where $d(\xi)$ denotes the Euclidean distance from an arbitrary point $\xi$ to its nearest path point $P_i \in \mathcal{P}$. By construction, $d(P_i)=0$ for all $P_i \in \mathcal{P}$, while $d(\xi)$ increases monotonically with the distance from the trajectory. 

We interpret this scalar function $d(\cdot)$ as generating a continuous field
\begin{equation}
U : \mathbb{R}^n \;\longrightarrow\; \mathbb{R}, \qquad U(\xi) = d(\xi),
\end{equation}
such that each point $\xi$ in the environment is assigned a value equal to its shortest Euclidean distance to the trajectory space $\mathcal{P}$. Intuitively, $U(\xi)$ measures how far $\xi$ is from the reference path and thus encodes a notion of proximity to the trajectory throughout the free space.  

To generate the guiding vector field on top of the distance field $U$, three steps are carried out: computing the inward normal direction, estimating the tangential direction, and combining them into a guidance law with distance shaping.

First, the normal direction is obtained from the gradient of the distance field. Since $\nabla U(\xi)$ points outward, we define the inward unit normal as
\begin{equation}
\boldsymbol{n}(\xi) = -\,\frac{\nabla U(\xi)}{\|\nabla U(\xi)\|}.
\end{equation}
In implementation, $\nabla U(\xi)$ is estimated through a local quadratic fitting procedure. Specifically, in a neighborhood of $\xi$ with local displacement $\delta=(\delta_x,\delta_y,\delta_z)$, the distance field is approximated as $U(\xi+\delta)\approx \theta_0+\theta_1\delta_x+\theta_2\delta_y+\theta_3\delta_z+\tfrac{1}{2}\theta_4\delta_x^2+\theta_5\delta_x\delta_y+\theta_6\delta_x\delta_z+\tfrac{1}{2}\theta_7\delta_y^2+\theta_8\delta_y\delta_z+\tfrac{1}{2}\theta_9\delta_z^2$, where the coefficients $\theta=(\theta_0,\dots,\theta_9)$ are obtained by least-squares regression over the sampled voxel values. Differentiating this polynomial yields a continuous approximation of the gradient, for example $\partial U/\partial x\approx \theta_1+\theta_4\delta_x+\theta_5\delta_y+\theta_6\delta_z$, with analogous expressions for $\partial U/\partial y$ and $\partial U/\partial z$. The resulting vector is finally normalized to obtain the inward unit normal $\boldsymbol{n}(\xi)=-\nabla U(\xi)/\|\nabla U(\xi)\|$.

Next, the tangential direction is estimated from nearby path points. Given a query position $\xi$, we first identify its closest path point $P_{i^\ast}=\arg\min_j \|\xi-P_j\|$. To ensure stability near boundaries, different local neighborhoods are selected: if $i^\ast=0$, we take the first three points $\{P_0,P_1,P_2\}$; if $i^\ast=n$ (the endpoint), we take the last three points $\{P_{n-2},P_{n-1},P_n\}$; otherwise we use the symmetric set $\{P_{i^\ast-1},P_{i^\ast},P_{i^\ast+1}\}$. From this neighborhood, the tangent direction is obtained by combining forward and backward chords. In particular, we define  
\begin{equation}
    \boldsymbol{\tau}(\xi) \;=\; 
    \frac{(P_{i^\ast+1}-P_{i^\ast}) + (P_{i^\ast}-P_{i^\ast-1})}
         {\big\|(P_{i^\ast+1}-P_{i^\ast}) + (P_{i^\ast}-P_{i^\ast-1})\big\|},
\end{equation}
which corresponds to the normalized average of local difference vectors and guarantees $\|\boldsymbol{\tau}(\xi)\|=1$. This expression can be interpreted as a finite-difference approximation of the local derivative of the path, obtained by combining forward and backward differences around $P_{i^\ast}$.

Finally, the normal and tangential directions are combined into the guiding vector field. Let $d(\xi)=U(\xi)$ and choose a convergence bandwidth $r>0$. We apply a shaping function $s: \mathbb{R} \to \mathbb{R}$
\begin{equation}
s(d(\xi)) = \tanh\!\Big(\frac{d(\xi)}{r}\Big),
\end{equation}
which smoothly saturates the effect of the distance. The resulting guiding vector field is then expressed as
\begin{equation}
\chi(\xi) = K_1\,\boldsymbol{\tau}(\xi) + K_2\,s\big(d(\xi)\big)\,\boldsymbol{n}(\xi),
\end{equation}
where $K_1,K_2>0$ are tuning parameters. Here the first term drives the robot forward along the path, while the second term ensures convergence back to the path when deviations occur. Fig.~\ref{f2} illustrates the resulting vector field structure, while Alg.~\ref{alg:gvf_construction} provides the corresponding computational procedure for constructing it. 
\begin{remark}
The discretization-based construction outlined above can approximate arbitrary trajectories and thus removes the reliance on implicit function representations that classical GVF methods require. Moreover, if the trajectory itself is generated in an online manner, for example through a receding-horizon planning (RHP) strategy, then the guiding vector field inherits this property and can be incrementally constructed in real time.
\end{remark}

\begin{algorithm}[t]
\caption{Incremental Guiding Vector Field Construction}
\label{alg:gvf_construction}
\begin{algorithmic}[1]
\State \textbf{Input:} Position $\xi$, discretized trajectory $\mathcal{P}$, ESDF $U$
\State \textbf{Parameters:} Convergence bandwidth $r$, gains $K_1, K_2$
\State \textbf{Output:} Guiding vector $\chi(\xi)$
\vspace{0.5em}
\State // \textbf{Normal direction from ESDF}
\State $d(\xi) \gets U(\xi)$
\State $g(\xi) \gets \nabla U(\xi)$ \Comment{via local quadratic fit}
\State $\mathbf{n}(\xi) \gets -g(\xi)/\|g(\xi)\|$
\vspace{0.5em}
\State // \textbf{Tangential direction from trajectory points}
\State $P_{i^\ast} \gets \arg\min_{P_j \in \mathcal{P}} \|\xi - P_j\|$ 
\State Select neighborhood $\{P_{i^\ast-1}, P_{i^\ast}, P_{i^\ast+1}\}$ (with boundary handling)
\State $\tilde{\tau} \gets (P_{i^\ast+1}-P_{i^\ast}) + (P_{i^\ast}-P_{i^\ast-1})$
\State $\boldsymbol{\tau}(\xi) \gets \tilde{\tau}/\|\tilde{\tau}\|$
\vspace{0.5em}
\State // \textbf{Distance shaping}
\State $s(\xi) \gets \tanh(d(\xi)/r)$
\vspace{0.5em}
\State // \textbf{Guiding vector field synthesis}
\State $\chi(\xi) \gets K_1\,\boldsymbol{\tau}(\xi) + K_2\,s(\xi)\,\mathbf{n}(\xi)$
\end{algorithmic}
\end{algorithm}
\subsection{Reference Trajectory Generation via Optimization}
\label{Reference Trajectory Generation via Optimization}
In the previous subsection, we described how an incremental guiding vector field can be constructed from a discretized trajectory space. To further improve this construction, a well-shaped reference trajectory is required to improve the robustness of the resulting vector field. For this purpose, we adopt a B-spline parameterization, which not only preserves the geometry of the original path but also provides a mathematically tractable form for subsequent optimization.

B-splines are particularly well-suited for reference trajectory generation because of their local support, smoothness, and convex hull property. Following previous work on spline-based planning~\cite{doi:10.1142/S230138502450033X,zhou2021raptor}, we parameterize the reference trajectory by a uniform clamped B-spline of degree $p$ with control points $\{\mathbf{C}_0,\dots,\mathbf{C}_N\}$ and knot vectors $\{u_0,\dots,u_{N+p+1}\}$. The resulting curve is written as 
\begin{equation}
\mathbf{p}(u)=\sum_{i=0}^N \mathbf{C}_i N_i^p(u) , u\in[u_p,u_{N+1}], 
\end{equation}
where $N_i^p(\cdot)$ are the B-spline basis functions.

To connect this curve with the discrete path $\mathcal{P}=\{P_0,P_1,\dots,P_M\}$, we assume that the path points are obtained by uniform sampling of the underlying continuous trajectory at times $\{t_m\}_{m=0}^M$. Formally, each point is defined as  
\begin{equation}
    P_m = \mathbf{p}\big(u(t_m)\big), \qquad m=0,\dots,M,
\end{equation}
where $t_m = t_0 + m\Delta t$ are uniformly spaced sampling times with interval $\Delta t$, and $u(t_m)$ denotes the corresponding B-spline parameter associated with time $t_m$. In this formulation, $M$ directly reflects the number of uniformly spaced samples along the trajectory, and the discrete path $\mathcal{P}$ is thus an explicit discretization of the continuous B-spline curve.
\begin{remark}
The accuracy of the guiding vector field depends on both the number of sampling points $M$ along the trajectory and the resolution of the discretized grid space. Higher grid resolution yields more precise numerical fields but increases computational and memory costs. Sparse sampling may lead to discontinuities in the reconstructed field, whereas overly dense sampling may cause multiple points to fall into the same grid cell. In our implementation, the grid resolution is not specifically optimized and is typically set to $0.1$. The core computational load lies in ESDF maintenance, which, based on the structure in~\cite{zhou2021raptor}, can be updated within milliseconds.
\end{remark}

We refine the reference trajectory by optimizing the interior control points $\{\mathbf{C}_p,\dots,\mathbf{C}_{N-p}\}$, keeping the boundary control points fixed to satisfy start and goal states. The optimization balances two objectives:
\begin{equation}
    \min_{\mathbf{C}} \; J_{\mathrm{total}} = \lambda_s J_s + \lambda_c J_c,
\end{equation}
where $J_s$ promotes smoothness of the spline and $J_c$ enforces obstacle avoidance with respect to the ESDF. Unlike model-based trajectory generation, we do not explicitly include higher-order dynamic feasibility terms (e.g., velocity, acceleration, jerk), since the trajectory is discretized again into a path space for GVF construction, and only the geometric properties are required.

\paragraph{Smoothness cost.}  
The smoothness objective penalizes variations in curvature by minimizing discrete approximations of higher-order differences between control points:
\begin{equation}
    J_s = \sum_{i=0}^{N-3} \|\mathbf{C}_{i+3} - 3\mathbf{C}_{i+2} + 3\mathbf{C}_{i+1} - \mathbf{C}_i\|_2^2.
\end{equation}
This formulation exploits the convex hull property of B-splines and ensures that the resulting curve remains globally smooth.

\paragraph{Collision cost.}  
For each interior control point $\mathbf{C}_i$, let $d(\mathbf{C}_i)$ denote its signed distance to the nearest obstacle from the ESDF, and let $d_{\mathrm{thr}}$ be a safety threshold. We penalize violations of the clearance condition using
\begin{equation}
\mathcal{F}(\mathbf{C}_i) =
\begin{cases}
    \big(d(\mathbf{C}_i) - d_{\mathrm{thr}}\big)^2, & d(\mathbf{C}_i) < d_{\mathrm{thr}}, \\
    0, & \text{otherwise},
\end{cases}
\end{equation}
and accumulate this penalty over all interior control points to form $J_c$. This encourages the spline to stay in obstacle-free regions while retaining flexibility to adapt its shape.

Through this construction, the optimized B-spline provides a collision-free and globally smooth trajectory that naturally fits into the guiding vector field framework. It serves as the geometric backbone for the GVF, ensuring that the vector field is anchored to a safe and well-structured path.
\begin{remark}
The optimization problem formulated above is unconstrained, and its gradient can be derived explicitly. This enables the use of quasi-Newton methods such as L-BFGS for efficient numerical solution. 
\end{remark}

\section{Results}
\label{Results}

\subsection{System Overview}
\subsubsection{Hardware Architecture}           
\begin{figure}[htbp]
    \centerline{
    \includegraphics[width=0.8\linewidth]{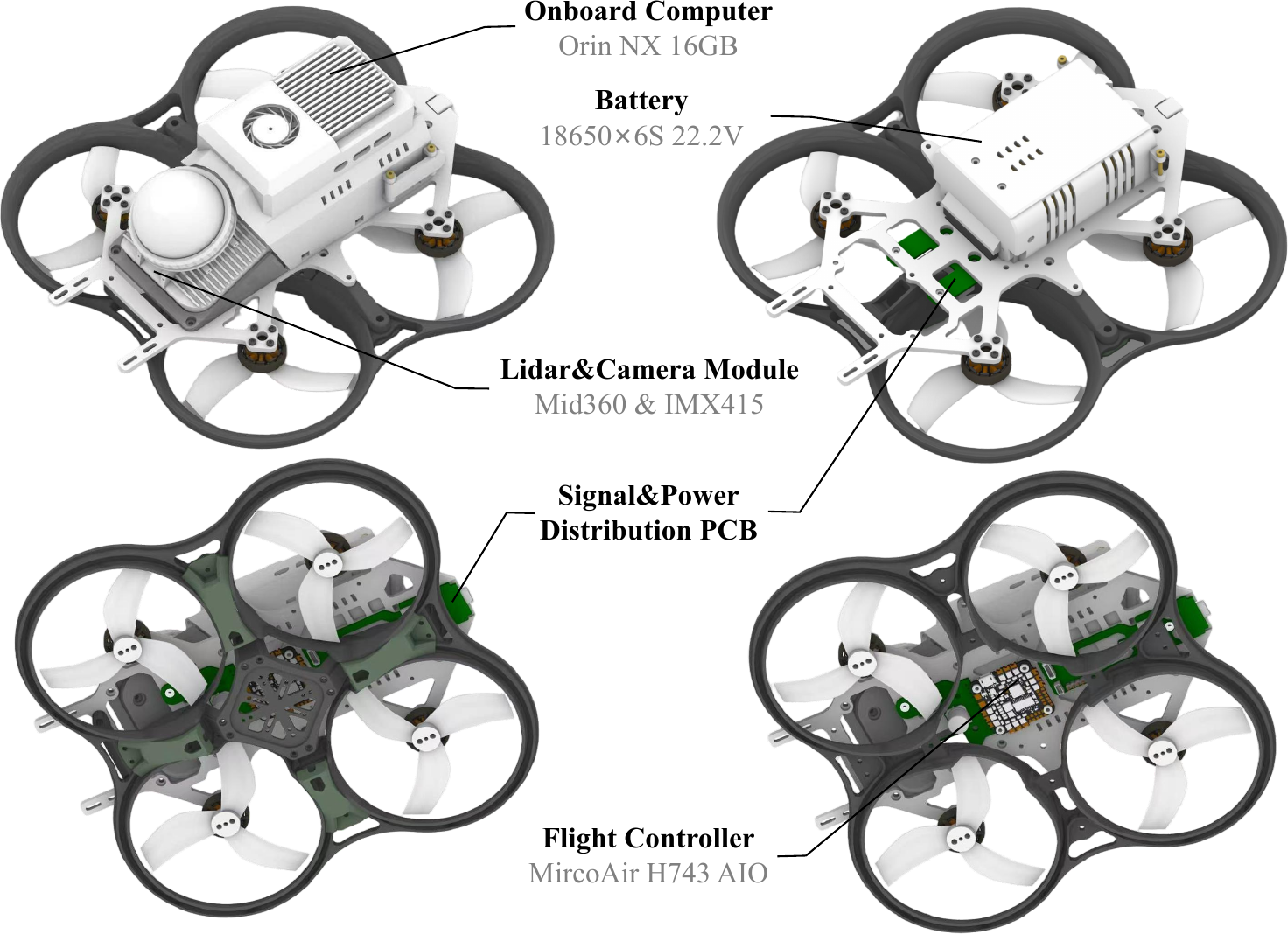}}
    \caption{
    Compact assembly and key component integration.
    }
    \label{f8}
\end{figure}

We present CU-Astro, a proprietary flight platform custom-developped on a 3.5-inch FPV quadrotor frame. Fig.~\ref{f8} illustrates the disassembled platform and its key components. The system integrates an onboard computer, a battery, a LiDAR and a camera module on the basic flight kit controlled by the MircoAir H743 AIO controller, which includes propellers, motors, electronic speed controllers (ESCs), and the airframe. A central proprietary PCB facilitates integrated signal transmission and power distribution from the battery. All components are assembled in a multi-layer stacked structure to maximize space efficiency and minimize total weight.

For the perception system, we developed a proprietary sensor module by redesigning the housing of the Livox Mid-360 LiDAR to allow direct camera mounting, which reduces the combined sensor weight by approximately 150 grams and thus extends flight endurance. The LiDAR is pitched downward by 22 degrees to enhance the ground-scanning perspective, thus improving localization robustness by capturing richer terrain features and increasing obstacle awareness during forward flight.

The airframe uses a carbon fiber structure with injection-molded propeller guards to suppress low-frequency vibrations and protect core components from collisions. Other structural parts are 3D-printed, with PETG used for the battery and computer enclosures, and 85A TPU for the LiDAR mount to dampen propeller-induced vibration and reduce IMU interference. The total weight of the platform is 649 g without the battery and 950 g with the battery, which consists of four 18650 cells supporting approximately 12 minutes of flight.

\begin{figure}[htbp]
\centerline{
\includegraphics[width=0.9\linewidth]{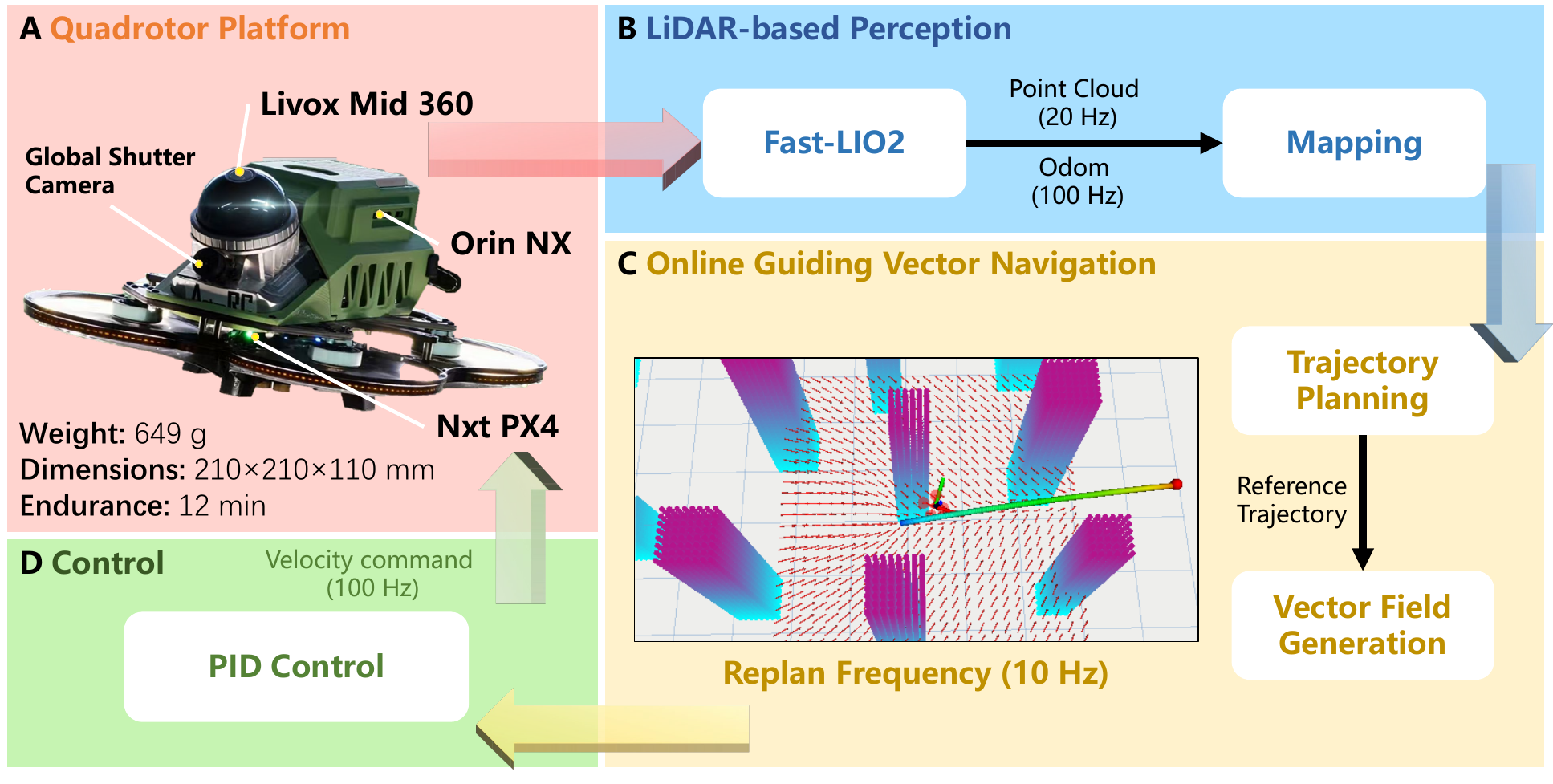}}
\caption{
System overview: \textbf{A}: Hardware description. \textbf{B}: Localization and mapping module. \textbf{C}: Navigation module via online guiding vector fields. \textbf{D}: Control module via cascaded PID controller.  
}
\label{f3}
\end{figure}    

\subsubsection{Software Architecture} 
The proposed navigation module, based on the Guiding Vector Field (GVF) planning method, seamlessly bridges the perception system and the flight control system. The overall architecture is depicted in Fig.~\ref{f3}. In this pipeline, FAST-LIO2 \cite{fast-lio2} is employed for real-time simultaneous localization and mapping. The perception module outputs point clouds at 20 Hz and odometry data at 100 Hz, which are then passed to the planning module. The planner generates a reference trajectory at 5 Hz according to the specified waypoints and converts it into corresponding control commands. These commands are sent to a cascaded PID controller, where the velocity loop operates at 100 Hz, ultimately generating motor control signals for the flight platform. Besides, to ensure precise timestamp synchronization between cloud points and inertial data, the IMU embedded within the Livox Mid-360 LiDAR is utilized instead of the flight controller's IMU. This integrated software system achieves centimeter-level precision in both localization and hovering stability, providing a reliable foundation for autonomous navigation.

\subsubsection{Initialization Parameters}
\label{System Initialization Parameters}
To facilitate reproducibility and ease the interpretation of our proposed method, we summarize the key control and planning parameters used throughout the paper in Tab.~\ref{SUMMARY OF VALUES OF METHOD PARAMETERS}. We recommend that readers consult the supplementary video for more detailed information regarding the content of this section.

\begin{table}[htbp]
\caption{VALUES OF PARAMETERS}
    \centering
    \begin{tabular}{cll}
\hline Parameter & Description & Values \\
\hline 
GVF & & \\
$K_1$ & Strength of aggregation & 1.5 \\
$K_2$ & Strength of dispersion & 1.5 \\
Planning & & \\
$T_p$ & Replanning interval & 0.2 s \\
$\lambda_s$ & Smoothness weight & 5 \\
$\lambda_c$ & Collision-avoidance weight & 10 \\
$d_{\text{thr}}$ & Obstacle safety threshold & 0.35 \\
\hline
    \end{tabular}
    \vspace{-0.7\baselineskip} % 
    \label{SUMMARY OF VALUES OF METHOD PARAMETERS}
\end{table}

\begin{figure*}[b]
\centerline{
\includegraphics[width=0.8\linewidth]{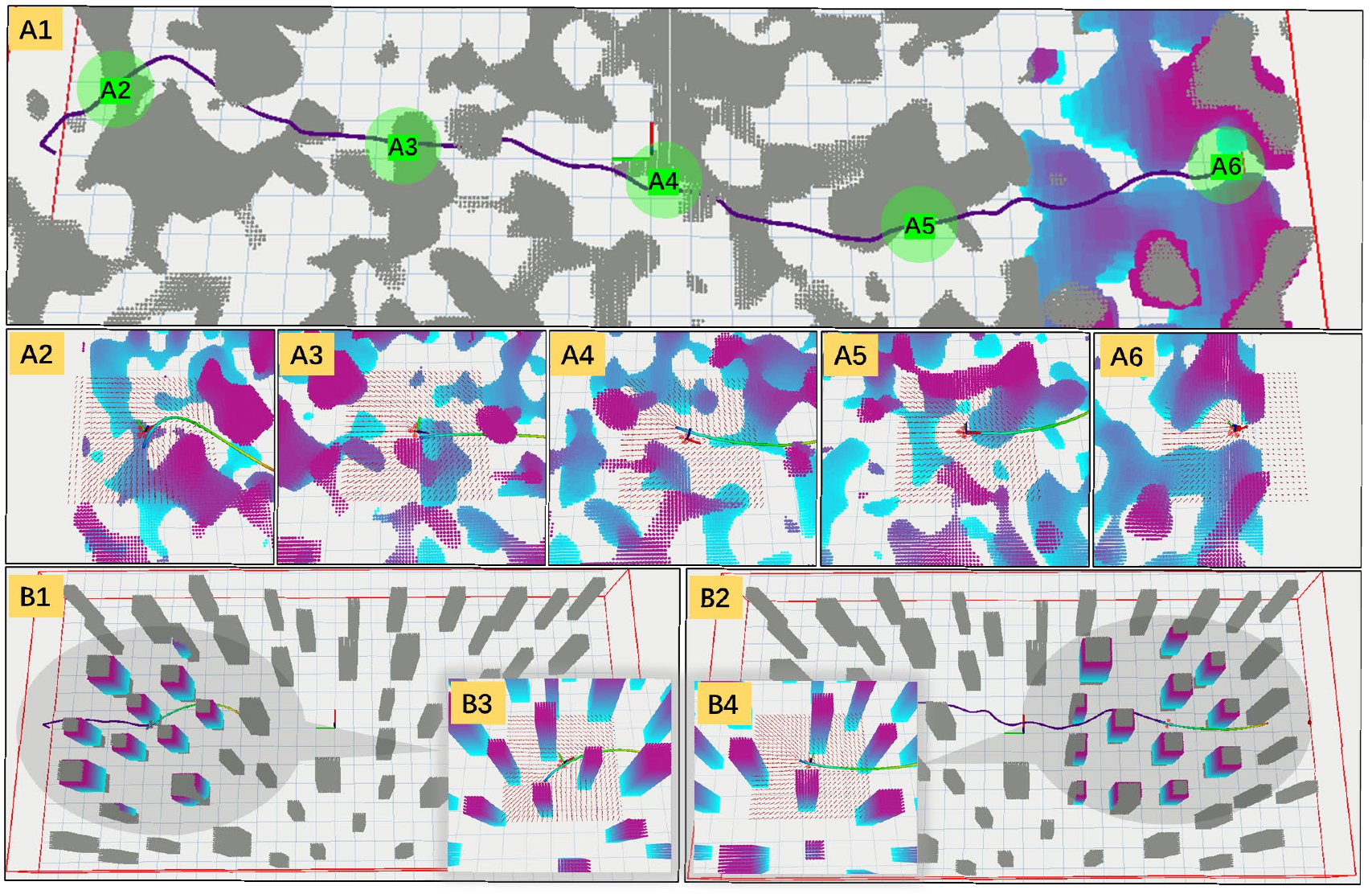}}
\caption{
Simulation experiments: 
\textbf{Scenario A}: Navigating dense, irregular obstacles.
\textbf{A1}: Overview of the flight path.
\textbf{A2-A6}: Sequential details of the online vector field-guided planning and avoidance process from start to destination.
\textbf{Scenario B}: Navigating a structured environment with rectangular pillars.
\textbf{B1 \& B2}: Overall flight path.
\textbf{B3 \& B4}: Zoomed-in views of quadrotor-pillar interactions.
}
\label{f6}
\end{figure*}

\subsection{Simulation Results}
\label{Simulation Results}
We conduct simulations using the Robot Operating System (ROS) on a workstation equipped with an Intel Core i9-14900KF processor running at 6.0 GHz and 64 GB of DDR5 memory, providing a total of 32 threads. The simulation leverages multi-threaded computation by parallelizing perception, planning, and control tasks across dedicated threads to maximize computational efficiency. The simulation environments are generated by mockamap \cite{mockamap} and comprise two distinct scenarios to validate the proposed method: Scenario A features a cluster of dense and irregular obstacles to evaluate 3D planning capability, while Scenario B consists of structured pillar-like obstacles to assess 2D planning performance, with specific results presented in Fig.~\ref{f6}. For better visualization clarity, the guiding vector field is depicted in 2D within the figure, while it is important to note that the actual planning process is conducted in 3D space using a 3D vector field.

The obstacle density in the environment is quantified as $d_{\text{obs}} = \frac{V_{\text{obs}}}{V_{\text{total}}}$, where $d_{\text{obs}}$ denotes the obstacle density, $V_{\text{obs}}$ is the total volume occupied by obstacles, and $V_{\text{total}}$ is the total volume of the environment.

In the experiments, the obstacle density is set to 50\% in Scenario A and 30\% in Scenario B. The experimental area is a $30\,\mathrm{m} \times 10\,\mathrm{m}$ environment, and the quadrotor maintains a cruise speed of 2 m/s while performing real-time obstacle perception, constructing an occupancy grid map, and maintaining a Euclidean signed distance field. It generates a B-spline curve as a reference trajectory to avoid all obstacles and constructs guiding vector fields online based on this B-spline to produce desired velocity commands for navigation. As shown in the results, the aircraft smoothly avoids all obstacles and successfully reaches the target from the start point.

\subsection{Real-world Experiments}
\label{Real-world Experiments}
To comprehensively validate the functionality and performance of the proposed method and system, real-world experiments were conducted in both indoor and outdoor environments. The following four experiments were designed to evaluate environmental adaptability in cluttered scenes and robustness against external disturbances. All experiments relied exclusively on the onboard perception module without any external positioning system, and with the quadrotor's cruise speed set at 1.5 m/s.

\subsubsection{Indoor Navigation in Dense Obstacles}
This experiment tests the method's ability to navigate through increasingly dense and irregular obstacle configurations indoors. The experimental area measures $30\,\mathrm{m} \times 8\,\mathrm{m}$, with an obstacle density of approximately 40\%. As shown in subfigures B1-B4 and C2 of Fig.~\ref{f7}, the vehicle successfully avoids obstacles while maintaining smooth motion, demonstrating reliable 3D obstacle awareness and collision-free path planning in confined spaces.

\subsubsection{Outdoor Navigation in Mixed-structured Environment}
\begin{figure}[htbp]
\centerline{
\includegraphics[width=0.87\linewidth]{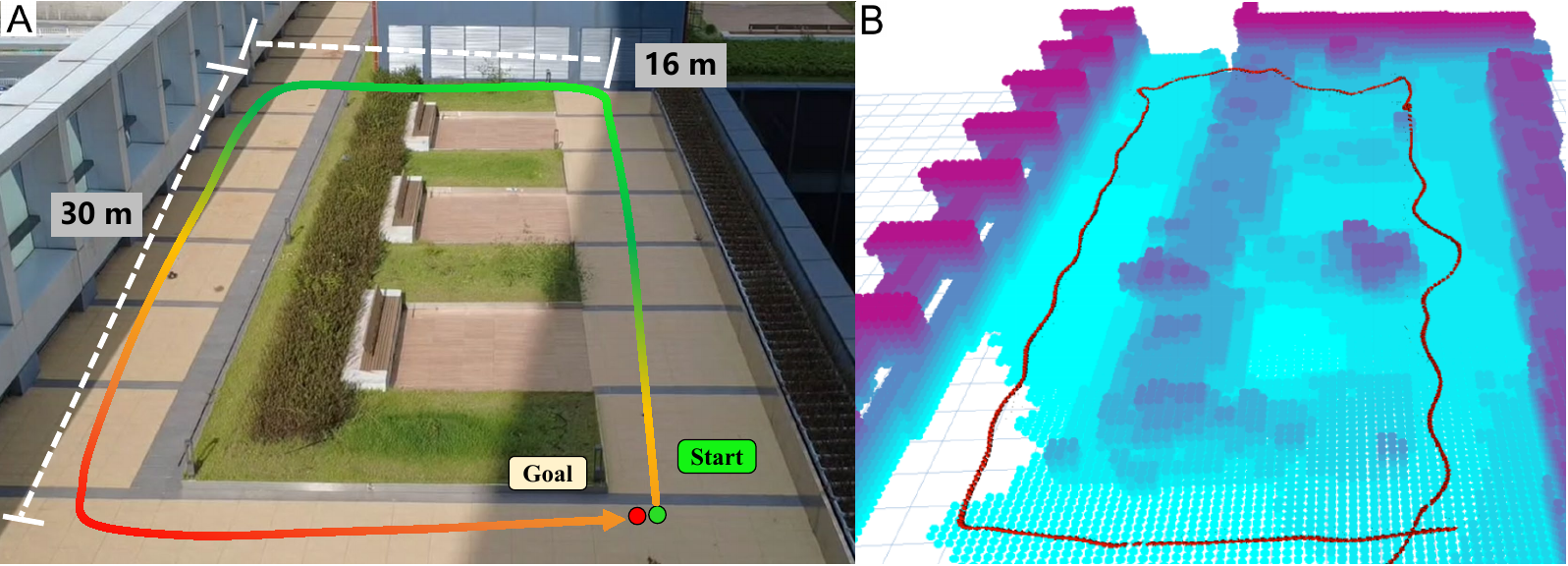}}
\caption{
Outdoor experiment results: 
\textbf{A}: Pre-defined quadrotor flight path: the quadrotor follows the perimeter of a rectangle and returns to the origin. 
\textbf{B}: The occupancy grid map and flight trajectory during the flight.
}
\label{f5}
\end{figure}

The system's ability to handle large-scale, mixed-structured scenes is verified in an outdoor setting. The experiment was conducted in a $30\,\mathrm{m} \times 16\,\mathrm{m}$ rooftop garden, providing an environment with both natural and artificial obstacles. Fig.~\ref{f5} shows the vehicle navigating through the terrain, where the method exhibits consistent performance without relying on prior maps or external infrastructure.

\subsubsection{Wind Disturbance Rejection Test}
\begin{figure}[htbp]
\centerline{
\includegraphics[width=0.9\linewidth]{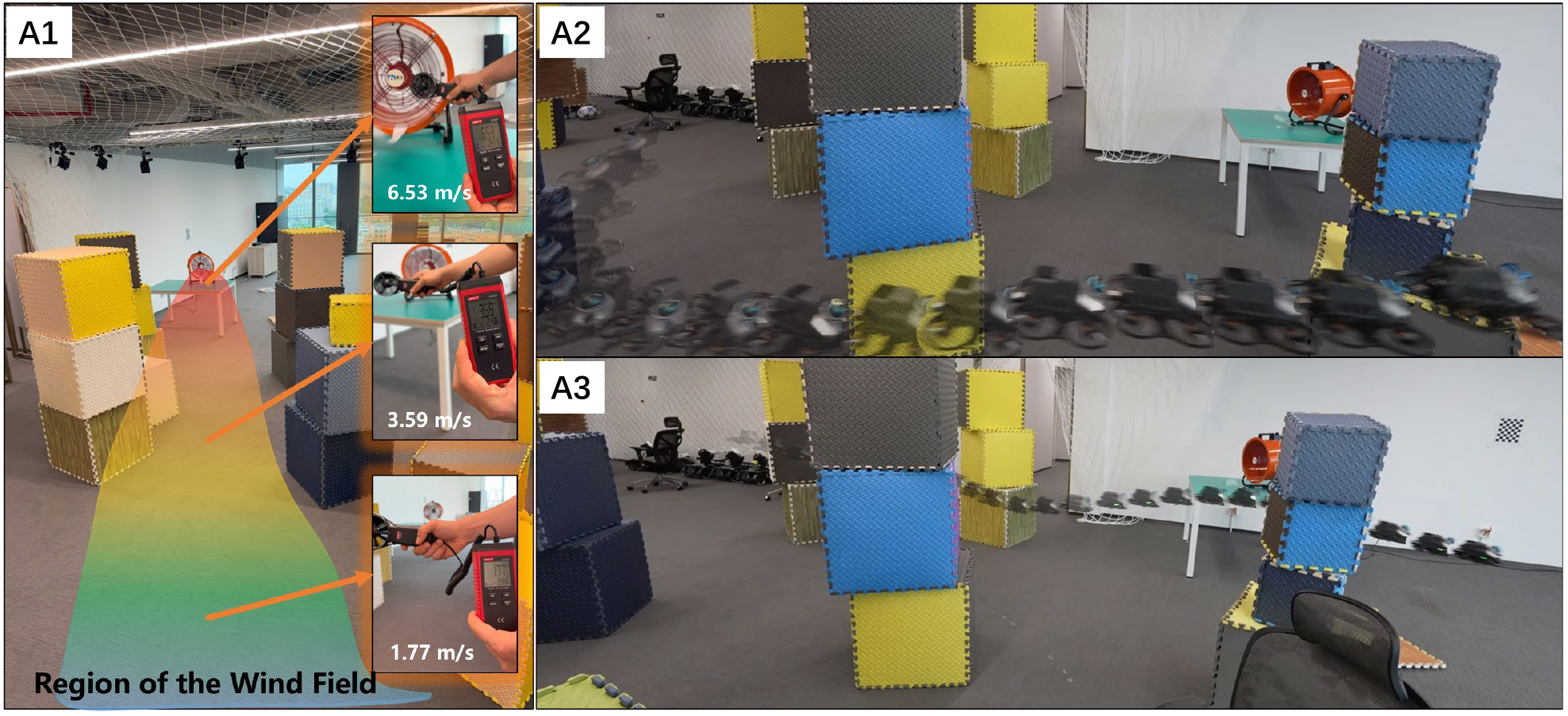}}
\caption{
Autonomous flight under wind field disturbances.
\textbf{A1}: Experiment setup showing wind speeds of 6.53 m/s, 3.59 m/s, and 1.77 m/s measured at distances of 10 cm, 1.5 m, and 3 m away from the fan.
\textbf{A2 \& A3}: The quadrotor's trajectory adapts in response to the varying wind speeds to maintain its target course.
}
\label{f4}
\end{figure}

To evaluate the closed-loop controller's robustness, an artificial wind field was applied to the aircraft during flight. The experiment setup, as shown in Fig.~\ref{f4}, involved varying wind speeds of 6.53~m/s, 3.59~m/s, and 1.77~m/s as the distance from the fan increased from 10~cm to 3.0~m.  Despite these lateral wind influences, the vehicle is able to successfully traverse the wind field and reach the designated waypoints under the guidance of the vector field.

\subsubsection{Manual Dragging Recovery Test}
\begin{figure*}[htbp]
\centerline{
\includegraphics[width=0.82\linewidth]{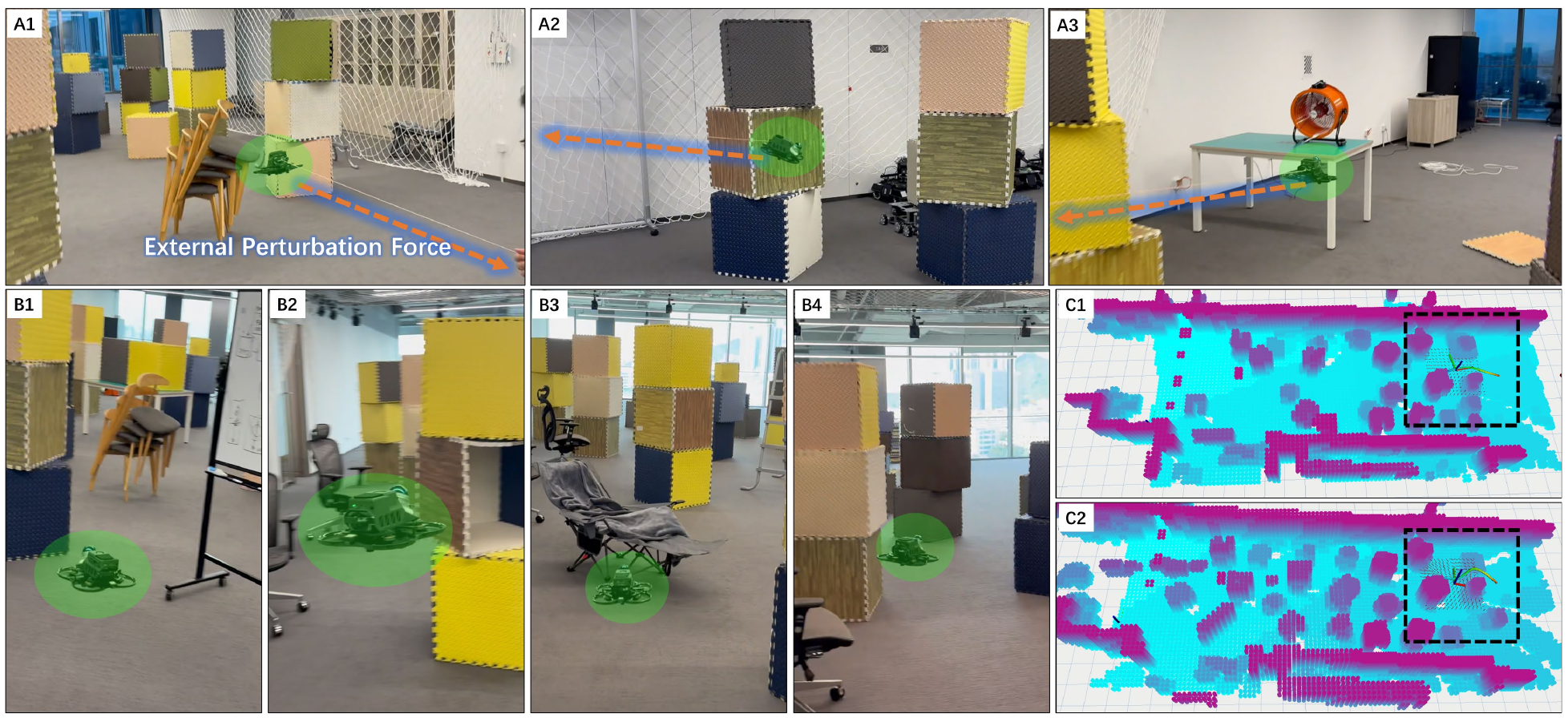}}
\caption{
Indoor experimental results: 
\textbf{A1-A3}: Quadrotor flight performance under imposed disturbances.
\textbf{B1-B4}: Autonomous obstacle avoidance in a dense environment.
\textbf{C1 \& C2}: Real-time incremental occupancy grid map construction and trajectory planning guided by vector fields for the scenarios in A and B, respectively.
}
\label{f7}
\end{figure*}

\begin{figure*}[htbp]
    \centerline{
    \includegraphics[width=0.83\linewidth]{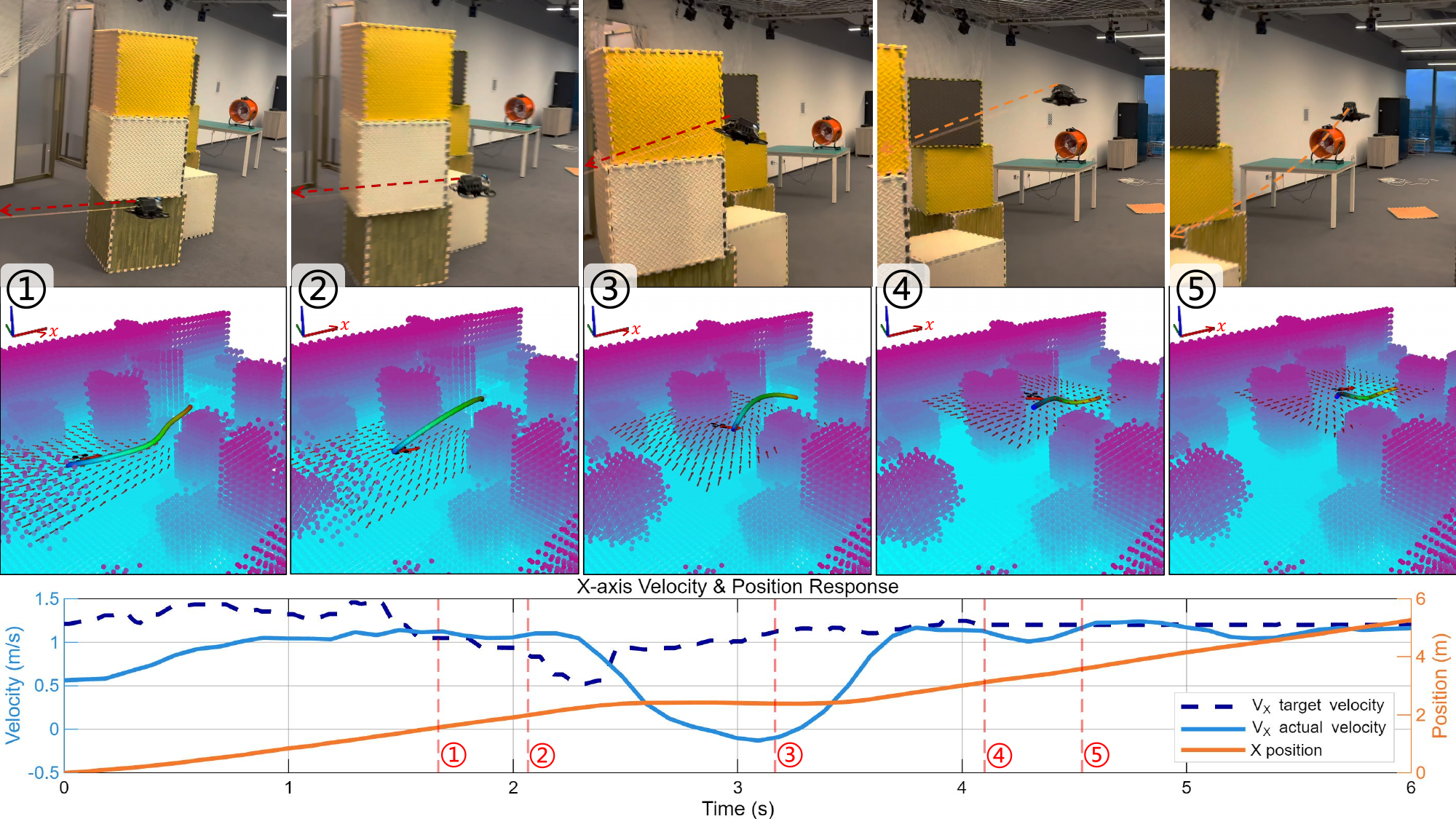}}
    \caption{
    Visualization of trajectory planning under imposed disturbances over a short period with position \& velocity profile.
    }
    \label{f9}
\end{figure*}

In this experiment, the aircraft was subjected to random manual dragging away from its planned trajectory to test the system's ability to recover from strong and unpredictable external disturbances, as shown in subfigures A1-A3 of Fig.~\ref{f7}. The direction and magnitude of the drag were not predetermined, simulating real-world random perturbations. During the experiment, the vehicle was forcibly dragged by hand, and under the influence of this external force, its velocity was reduced to zero, causing it to come to a complete stop at an unintended position away from the planned path. Throughout the process, the system continuously perceived the disturbed state and updated the guiding vector field in real time. Once the manual drag was released, online guiding vector field allowed the aircraft to autonomously reorient and smoothly return to the original trajectory. Fig.~\ref{f9} captures five key timestamps before, during, and after the disturbance, illustrating the process of disturbance, forced stop, and subsequent recovery.

\subsection{Comparison with Benchmarks}
\label{Comparison with Benchmarks in Complex Obstacle Scenarios}
We conducted a comparative study against several representative local navigation methods, and the results are summarized in Tab.~\ref{tab:comparison_single}. All tests and comparisons were carried out in simulation environments. For each scenario, 100 independent trials were performed, and the average planning time, success rate, travel time, and travel distance were recorded for statistical evaluation. The primary focus was on evaluating the real-time performance of each algorithm. All the compared methods are capable of maintaining satisfactory real-time operation in complex environments, indicating their effectiveness for navigation in unknown environments.

\begin{table}[htbp]
\caption{Method Comparison Results in Simulation}
\label{tab:comparison_single}
\resizebox{0.47\textwidth}{!}{
\centering
\footnotesize
\setlength{\tabcolsep}{3.5pt}
\renewcommand{\arraystretch}{1.05}
\begin{tabular}{cccccc}
\hline
\hline
\textbf{Scene} & \textbf{Method} & \textbf{\begin{tabular}[c]{@{}c@{}}Avg.\\ Planning\\ Time (ms)\end{tabular}} & \textbf{\begin{tabular}[c]{@{}c@{}}Success\\Rate (*/100)\end{tabular}} & \textbf{\begin{tabular}[c]{@{}c@{}}Avg.\\ Travel\\ Time (s)\end{tabular}} & \textbf{\begin{tabular}[c]{@{}c@{}}Avg.\\ Dist.\\ (m)\end{tabular}} \\ 
\hline
\multirow{4}{*}{\textbf{\begin{tabular}[c]{@{}c@{}}No External\\ Disturbance\\ (Scenario A)\end{tabular}}}
       & Fast Planner\cite{zhou2021raptor} & 7.14 $\pm$ 1.25 & 100\% & 62.3 $\pm$ 6.8 & 75.4 $\pm$ 5.7 \\
& \textbf{Ego Planner\cite{zhou2020ego}} & \textbf{5.73} $\pm$ \textbf{1.09} & 100\% & \textbf{61.1} $\pm$ \textbf{5.9} & \textbf{72.7} $\pm$ \textbf{4.9} \\
& CCNMPC\cite{zhu2019chance} & 14.60 $\pm$ 2.88 & 72\% & 84.1 $\pm$ 9.5 & 79.4 $\pm$ 6.2 \\
& GVF (Ours) & 9.95 $\pm$ 1.67& 100\% & 68.9 $\pm$ 7.1 & 77.3 $\pm$ 5.4 \\
\hline
\multirow{4}{*}{\textbf{\begin{tabular}[c]{@{}c@{}}With External\\ Disturbance\\ (Scenario A)\end{tabular}}}
& Fast Planner\cite{zhou2021raptor} & 7.21 $\pm$ 1.35 & 24\% & 73.5 $\pm$ 8.2 & 79.8 $\pm$ 6.3 \\
& Ego Planner\cite{zhou2020ego} & \textbf{5.84} $\pm$ \textbf{1.10} & 22\% & 72.1 $\pm$ 7.0 & 75.2 $\pm$ 5.2 \\
& CCNMPC\cite{zhu2019chance} & 14.78 $\pm$ 2.95 & 12\% & 95.7 $\pm$ 10.3 & 85.1 $\pm$ 6.8 \\
& \textbf{GVF (Ours)} & 9.90 $\pm$1.67 & \textbf{92\%} & \textbf{69.5} $\pm$ \textbf{6.8} & \textbf{73.8} $\pm$ \textbf{5.0} \\
\hline
\hline
\end{tabular}% 
}
\end{table}

In terms of implementation, Fast Planner~\cite{zhou2021raptor} employs an Euclidean Signed Distance Field to ensure trajectory safety, while Ego Planner~\cite{zhou2020ego} computes environmental gradients directly in the grid map, which leads to superior computational efficiency. CCNMPC~\cite{zhu2019chance}, on the other hand, formulates the problem as a constrained MPC within a convex optimization framework, resulting in higher computational overhead. Correspondingly, our proposed GVF-based method maintains a local guiding vector field along the reference trajectory, which slightly increases computation time compared to~\cite{zhou2021raptor} and~\cite{zhou2020ego}. Nevertheless, it should be noted that the additional computation mainly arises from maintaining the gradient field, which is inherently efficient. Therefore, the overall computation remains well within real-time requirements, even in unknown and complex environments.

In the disturbance experiments, external perturbations were artificially introduced by applying additional forces through manual keyboard input. Specifically, a user-generated virtual force command was superimposed on the quadrotor's control inputs during flight, temporarily pushing it away from the planned trajectory to simulate unexpected disturbances, with the maximum induced acceleration reaching up to 1.5~m/s\textsuperscript{2}. In these experiments, the closed-loop property of the guiding vector field demonstrated a significant advantage. When external disturbances caused the quadrotor to deviate from its nominal trajectory, the navigation success rates of other methods dropped sharply: the quadrotor often could not return to the planned path and failed to reach the target safely. These methods struggled to compensate for such deviations, resulting in frequent collisions or navigation failures, which reflects the high sensitivity of open-loop schemes to perturbations as discussed earlier. In contrast, the vector field-based method generates navigation commands directly from the current state, maintaining an inherent closed-loop behavior. Even after being disturbed away from the planned trajectory, the system could reliably guide the quadrotor back to the reference path and continue navigation. Therefore, the proposed vector field framework significantly enhances robustness to unknown environmental disturbances and maintains satisfactory real-time performance, even when other methods exhibit markedly reduced success rates under similar disturbance conditions.

\subsection{Discussion}
\label{Discussion}
Based on the above experiments, it can be observed that in unknown environments, the performance of the proposed real-time vector field navigation method is almost comparable to that of trajectory planning approaches. However, when external disturbances are introduced into the environment, the vector field exhibits a clear advantage in disturbance rejection. This observation supports the interpretation of the vector field method as a form of \emph{closed-loop} trajectory planning. It is worth noting that during the construction of the guiding trajectory for the vector field, we still formulate a multi-objective optimization problem to generate the reference trajectory, which is not fundamentally different from several mainstream frameworks. Consequently, if trajectory planning techniques continue to advance, the proposed vector field approach will also benefit from such improvements.

Compared with some traditional vector field methods, our approach does not require constructing the intersection of multiple implicit function surfaces to generate the vector field. Instead, it relies on the ESDF generated from the discretized trajectory space. This greatly reduces the difficulty of designing vector fields. In real-world navigation scenarios, reference trajectories inevitably exhibit complex shapes, making it challenging to directly approximate them using multiple implicit functions, which significantly limits the feasible operational space. In practice, many of these traditional approaches are effective only in open environments or in scenarios where obstacles are known in advance, but they are not effective in complex and unknown environments. Therefore, the proposed incremental vector field method constitutes a substantial improvement.

\section{Conclusion and Future Work}
\label{Conclusion and Future Work}
In this paper, we proposed an online incrementally constructed guiding vector field for real-time navigation in unknown complex environments. The proposed method generates reference guiding trajectories in real time and discretizes the trajectory space to construct the vector field. The normal guiding component is derived from the Euclidean signed distance field constructed in the discretized space, while the tangential component is obtained by polynomial fitting of the path. The entire vector field can be generated online at 20 Hz on edge computing devices, enabling quadrotors to autonomously navigate in unknown complex environments while remaining robust against unforeseen disturbances such as wind or human intervention. The system has been extensively validated in both simulations and physical experiments, and the results demonstrate that the proposed approach achieves remarkable robustness and real-time performance.

However, the proposed method still has certain limitations. The current vector field is designed using tangential and normal components, which cannot fully meet the requirement of high tracking accuracy. Although the convergence of the vector field can be theoretically guaranteed, the present design does not ensure bounded convergence time, which is critical in time-sensitive scenarios. Recent studies have explored the use of finite-time guiding vector fields to explicitly guarantee finite-time convergence in such applications. Nevertheless, these approaches are still under development, and further research is needed to address their limitations and fully integrate them into practical systems. In addition, we plan to extend the proposed framework to underwater robotic platforms, since the closed-loop, model-free nature of our method has the potential to handle the complex, strongly coupled hydrodynamics that are difficult to model accurately in underwater scenarios.

% References
\footnotesize
\bibliographystyle{IEEEtran}
\bibliography{cite}
\enlargethispage{-\baselineskip}

\end{document}